\documentclass{article}
\usepackage{spconf,mystyle,graphicx,epstopdf}

\def\A{{\mathbf A}}
\def\b{{\mathbf {b}}}

\title{Geometry of Deep Learning for Magnetic Resonance Fingerprinting }

\name{Mohammad Golbabaee$^1$, Dongdong Chen$^2$, Pedro A. G{\'o}mez$^{3,4}$, Marion I. Menzel $^4$ and Mike E. Davies$^{2}$.\vspace{-.35cm}\thanks{This work is funded
		by the EPSRC grant EP/M019802/1, ERC C-SENSE project (ERCADG-2015-694888), and the Scottish Research Partnership in engineering (SPRe) Award PECRE1718/18.}}

	\address{\small{$^1$Computer Science Department, University of Bath, $^2$School of Engineering, Universty of Edinburgh.}\\ \small{ $^3$Munich School of Bioengineering, Technische Universit{\"a}t M{\"u}nchen.\
	$^4$GE Healthcare, Munich. } \vspace{-.1cm} }
\begin{document}
\ninept
\maketitle
\begin{abstract} 
	Current popular methods for Magnetic Resonance Fingerprint (MRF) recovery are bottlenecked by the heavy storage and computation requirements of a dictionary-matching (DM)
	step due to the growing size and complexity of the fingerprint dictionaries in multi-parametric quantitative MRI applications. In this paper we study a deep learning approach 
	to address these shortcomings. Coupled with a dimensionality reduction first layer,  the proposed MRF-Net is able to reconstruct quantitative maps by saving more than 60 times in memory and computations required for a DM baseline.
	Fine-grid manifold enumeration i.e. the MRF dictionary is only used for training the network and not during image reconstruction. We show that the MRF-Net provides a piece-wise affine approximation to the Bloch response manifold projection and that rather than memorizing the dictionary, the network efficiently clusters this manifold and learns a set of hierarchical \emph{matched-filters} for affine regression of the NMR characteristics in each segment.


\end{abstract}

\section{Introduction}
\label{sec:introduction}


Magnetic Resonance Fingerprinting (MRF) recently emerged to accelerate acquisition of the  \emph{quantitative} NMR characteristics such as 
the T1, T2 and T2$^*$ relaxation times, field inhomogeneity and perfusion~\cite{MRF,FISP, MRF-perfusion,EPIT1T2star}. 
As opposed to mainstream qualitative assessments these \emph{absolute} physical  quantities can be used for
tissue or pathology identification independent of the scanner or scanning sequences. Unlike conventional quantitative approaches MRF uses  i) short and often complicated excitation pulses which encode many NMR parameters simultaneously, and ii) significantly undersampled k-space data. 
To overcome the lack of sufficient spatio-temporal information MRF incorporates a physical model based on exhaustively simulating a large dictionary of magnetic responses (fingerprints) for all combinations of the quantized NMR parameters. This dictionary is then used for matched-filtering in a model-based reconstruction scheme e.g.~\cite{BLIPsiam}.  
As occurs to any multi-parametric manifold enumeration, the main drawback of such approach is the size of  this dictionary which grows exponentially in terms of the number of parameters and their quantization resolution;  a serious (non-scalability) limitation of the current methods to be applicable in the emerging multi-parametric MRF applications. 
In conjunction with the widespread applications of machine learning methodologies, 
a number of recent empirical studies have proposed a dictionary-free deep learning (DL) approach to address this shortcoming~~\cite{DRONE-MRF,betterreal,hoppeDLMRF,balsigerMRFDL} the crux of which is to bypass the DM step by using  compact deep neural networks. 
However, reasons that DL works so well for this problem are poorly understood. 

\begin{figure}[t!]
	\centering
	\subfloat
	{\includegraphics[width=\linewidth]{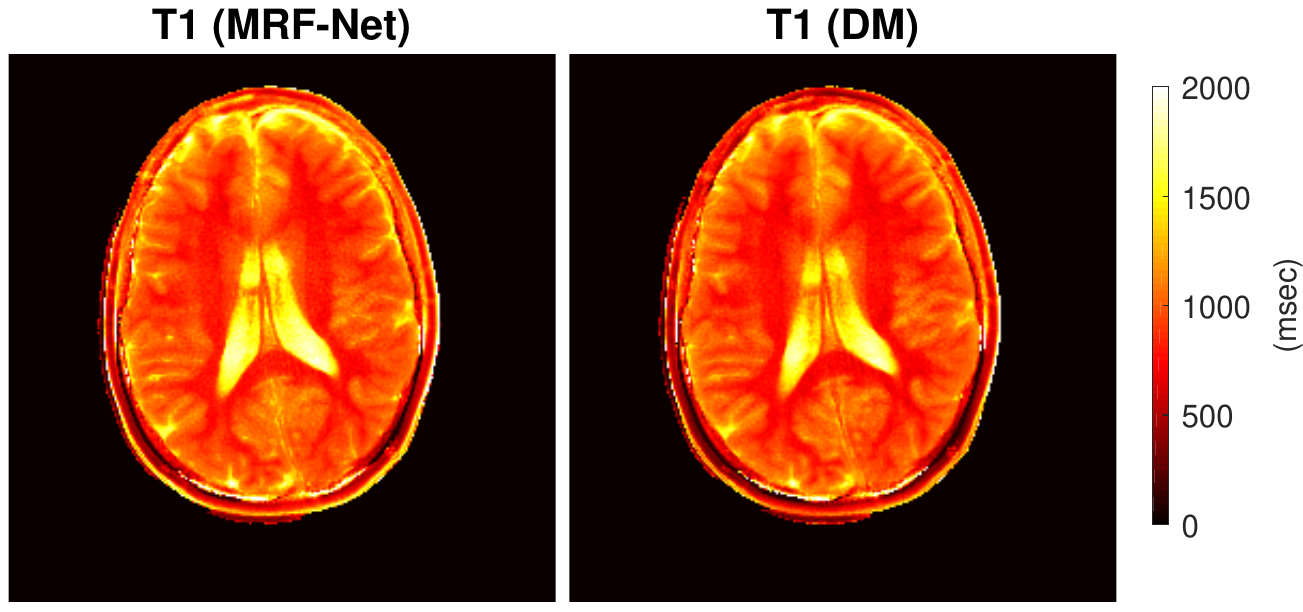}}
	\\
	\subfloat
	{\includegraphics[width=\linewidth]{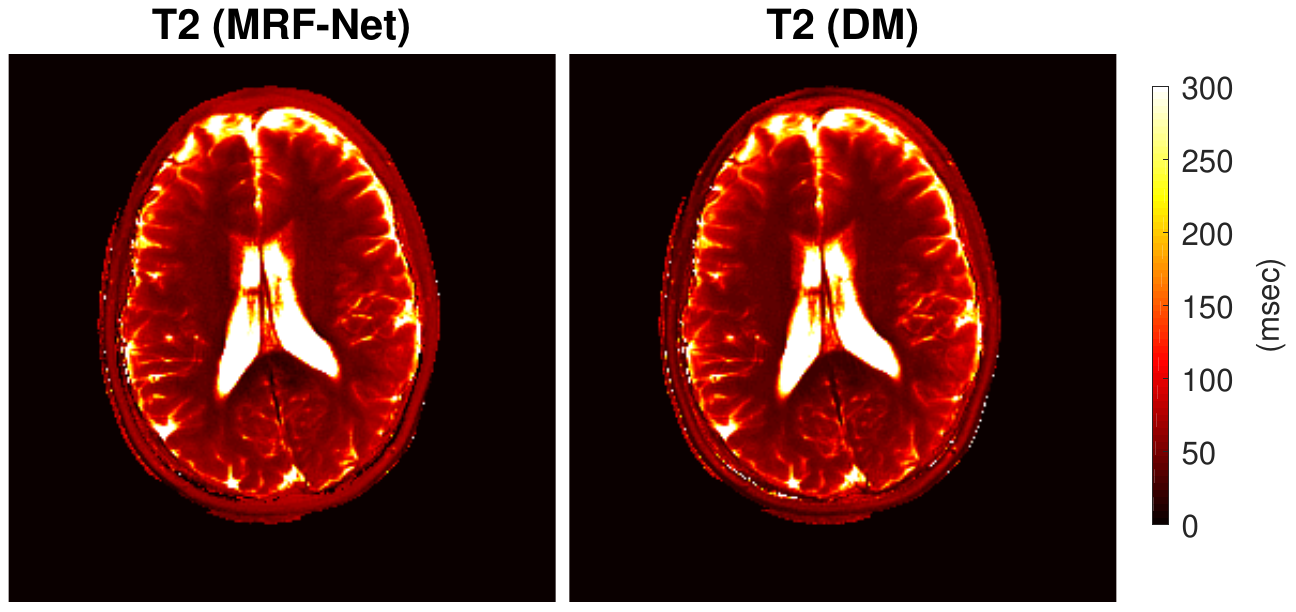}}
	\caption{Reconstructed T1 and T2 maps using the proposed dictionary-less MRF-Net and dictionary matching (DM) baseline.}
	\label{fig:maps}
\end{figure}

This paper aims at uncovering the underlying mechanisms by which DL achieves such progress for the MRF framework from a geometrical point of view. 
We show that the MRF-Net provides a piece-wise affine approximation to the Bloch response manifold projection and that rather than memorizing the dictionary, the network efficiently clusters this manifold layer-by-layer and implicitly learns a set of hierarchical \emph{matched-filters} for affine parameter regression in each segment.
Besides, we propose a competitive architecture to the current DL baselines. The proposed MRF-Net features a (unsupervisedly learned) dimensionality reduction first layer which promotes a low-rank subspace prior during image reconstruction, and results in less units and training resources as required for the uncompressed DL approaches earlier proposed for this problem.  Our \emph{in-vivo} experiment for estimating two NMR quantities (i.e. a small-size MRF problem) indicates that the MRF-Net is capable of saving more than 60x in storage and model-fitting computations as required for a dimension-reduced DM baseline (Figure~\ref{fig:maps}).



\vspace{-.25cm}
\section{Problem statement}
\vspace{-.25cm}
\label{sec:model}
%

MRF acquisitions follow a linear spatio-temporal model:
\eql{\label{eq:forward}
	Y\approx \Aa (X),
}
where $Y\in \CC^{m\times L}$ denotes noisy k-space measurements collected at $t= 1,\hdots,L$ temporal frames after each  excitation. 
The MRF image sequence is a complex-valued matrix $X$ of spatio-temporal resolution $n\times L$  i.e. $n$ spatial voxels and $L$ temporal frames. The forward operator 
$\Aa:=P_\Omega F S(.)$
models multi-coil sensitivity maps $S$ 
and a sub-sampled Fourier operator $P_\Omega F$ which represents the k-space acquisition with respect to a set of \emph{temporally-varying} locations $\Omega=\bigcup_{t=1}^L\Omega_t$ where $\Card(\Omega_t)=m\ll n$.

The main source of quantitative measurements are the per-voxel magnetization response of proton dipoles obtained from dynamic rotations of the external magnetic field i.e. a sequence of Flip Angles (FA) $\{\alpha_t\}_{t=1}^L$ applying at certain repetition times $\{TR_t\}$.\emph{ Tissues with different NMR characteristics respond distinctively to these excitations.} The MRF framework relies on this principle to regularize the under-determined problem~\eqref{eq:forward} by a temporal model and enable parameter estimation. Magnetization trajectories (responses) \textemdash denoted by $\Bb(\Theta; TR,\alpha)\in \CC^L$ \textemdash  are distinct solutions of the  \emph{Bloch differential equations} 
for a given set of intrinsic NMR parameters $\Theta\in \RR^P$ and excitation sequence $\{\alpha_t,TR_t\}$~\cite{jaynes1955matrix}. 
Current MRF approaches discretize through a dense sampling the parameter space $[\Theta]=[T1]\times[T2]\times \ldots$
and simulate a large dictionary of normalized fingerprints $D=\{D_j\}_{j=1}^d$ 
where, 
\eql{\label{eq:fingerprints} D_j := \Bb([\Theta_j]; TR,\alpha)\quad \forall j=1,\hdots,d,} 
for all $d$ combinations of the quantized parameters. 
Under the \emph{voxel purity} assumption  
each spatial voxel of the MRF image corresponds to a unique NMR parameter and would approximately match to a temporal trajectory in the fingerprint dictionary: $X_v \in D\quad  \forall v=1,\hdots, n,$
where $X_v$ denotes the normalized $v$-th row of $X$ i.e. a multi-dimensional spatial voxel.


\section{Parameter estimation}
\label{sec:second-section}

A popular approach for parameter estimation is to perform  back-projection (adjoint operator) on the k-space data $\widehat X :=\Aa^H(Y)\in\CC^{n\times L}$ followed by dictionary matching  to identify the highest correlated atom and its corresponding NMR parameters for each (normalized) voxel of the highly aliased back-projected image $\widehat X$:
\eql{\label{eq:NNS}
	[\Theta_v] = \text{NNS}_{D}(\widehat X_v), \quad \forall v=1,\hdots,n.
}
Here $\text{NNS}_D(x):=\argmin_j\norm{x-D_j}_2$ denotes the nearest neighbour search 
which serves as a Euclidean projection onto the discrete set of fingerprints i.e. the manifold of Bloch Eq solutions. 
A temporal (subspace) compression can
be used to shrink the search dimension i.e. $\widetilde X_v:=V_s^H \widehat X_v,\widetilde D_j := V_s^HD_j$ across the $s\leq L$ dominant principal components
of $DD^H \approx V_s \Lambda V_s^H$~\cite{SVDMRF}, and to promote a low-rank subspace prior during image reconstruction~\cite{asslanderLR,zhaoLR}. However, enumerating the multi-parametric MRF manifold in order for~\eqref{eq:NNS} to be an accurate projection introduces an exponentially growing complexity (in terms of $P$) to the storage and computations needed for conducting NNS. 
A recent line of research~\cite{CoverBLIP-MLSP, inexactipg-tit} shows that certain tree search strategies can benefit from 
the low \emph{intrinsic dimensionality} of the MRF manifold and significantly accelerate the matching step. 
However storage of the dictionary or the corresponding tree still remains a big challenge for fine-grid enumerations.

\subsection{MRF-Net}
\begin{figure}[t!]
	\centering
	\includegraphics[width=.85\linewidth]{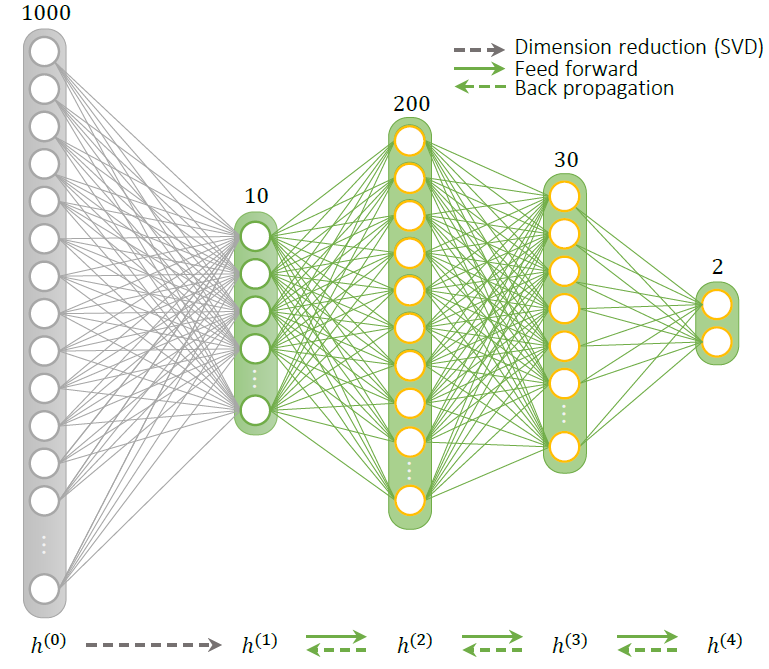}
	\caption{Illustration of the MRF-Net: pre-trained linear dimension reduction used for the first layer. Three last layers use ReLU non-linearities (orange) and are trained by the standard back-propagation.  }
	\label{fig:training}
\end{figure}

In this study we propose training 
a 4-layer fully connected feed-forward network dubbed as the MRF-Net for
approximating the MRF manifold projection by a continuous mapping $\Ff:\CC^L\rightarrow \RR^p$:
\eql{\Theta_v = \Ff(\widehat X_v),
}
where $\Ff(x) \approx \argmin_{\Theta} \norm{x-\Bb(\Theta)}_2$. 
The first layer of MRF-Net unsupervisedly learns the best linear projection onto the subspace of clean fingerprints through principal component analysis, and it is kept fixed during training other layers. 
 Three other layers include nonlinear ReLU activations in order to  approximate the dimension-reduced projection function $\argmin_{\Theta} \norm{V_s^Hx-V_s^H\Bb(\Theta)}_2$. 
The size of MRF-Net is $1000-10-200-30-2$ including an input layer $L=1000$ fed with voxel sequences form the  back-projected images, and 4 hidden layers as shown in Figure~\ref{fig:training}. Dimensions of the input/output and hidden units are customized here for the Steady State
Precession (FISP) sequence~\cite{FISP} which encodes $P=2$ NMR characteristics i.e. $\Theta=\{T1,T2\}$ relaxation times. 
The MRF dictionary corresponding to the FISP sequence is shown to be well represented by very few  principal components~\cite{SVDMRF} e.g. $s=10$ here, which determines first layer's dimension accordingly. 
Thanks to this dimensionality-reduction, MRF-Net requires far less units and training resources compared to 
the uncompressed DL approaches
proposed earlier in~\cite{DRONE-MRF,betterreal}. 

\vspace{-.25cm}
\subsubsection{Training MRF-Net}\label{sec:training}
\vspace{-.15cm}
Fine-grid manifold enumeration i.e. the MRF dictionary is only used for training and not during image reconstruction.  
To avoid loosing discrimination between fingerprints \textemdash by the magnitude-only data treatment proposed in~\cite{DRONE-MRF} \textemdash we adopt a phase-alignment heuristic used in practice~\cite{coverblip-journal,AIRMRF}
to align dictionary atoms (for training) and back-projected images (the input). With this treatment we can assume that without loss of generality the MRF-Net consists of real-valued parameters and approximates a real-valued mapping.

A fully connected feed-forward  network is  composed of $N$ (here $N=4$) layers each applying an affine transform followed by non-linear activation functions. The network can be modelled as $\mathcal F\equiv h^{(N)}(x):\RR^L\rightarrow \RR^P$ through a hierarchy of hidden mappings $h^{(i)}(x):\RR^L\rightarrow \RR^{\dim(i)}$ initialized by inputs $h^{(0)}(x)=x$:
\begin{align}
&h^{(i)}(x) =f_i\left(W^{(i)}h^{(i-1)}(x) + \beta^{(i)}\right) \, \text{for} \quad i=1,2,\ldots,N. \label{eq:layers}
\end{align}		
Here $W^{(i)}\in \RR^{\dim(i)\times \dim(i-1)}$ and  $\beta^{(i)}\in\RR^{\dim(i)}$ are the weights and biases at the $i$-th layer and $f_i(\cdot)$ is an element-wise nonlinear activation function. The first Layer of MRF-Net consists of identity activation $f_1(u)=u$, and unsupervised trained parameters $W^{(1)}=V_s^H, \beta^{(1)}=\mathbf{0}$. 
The remaining layers use ReLU activations $f_i(u)=\max(u,0)$ for $i=2,3,4$.  To learn these layers, MRF-Net is supervised trained to minimize the mean-squared regression loss between $h^{(4)}_k,\Theta_k$.\footnote{We use Adam optimizer~\cite{adam} where the gradient updates 
	are computed by the standard back-propagation on a moderate CPU desktop. Optimization parameters are as follows: batch size 50, 30 epochs and the step-size $10^{-2}$ decaying at the rate of $0.8$ after each epoch.}  
Training inputs are dimension-reduced (phase-aligned) atoms of the fine-grid MRF dictionary $\widetilde D_k$ corrupted by zero-mean independent Gaussian noises $\xi_k$ with SNRs randomly selected between 40-60 dB. We use the Extended Phase Graph framework~\cite{EPGWeigel} to simulate Bloch Eq responses to the FISP sequence for all combinations of $T1$=[100:10:4000] (msec) and $T2$=[20:2:600] (msec), and build a dictionary with $d=113781$ atoms for training. 
After noise corruption (i.e. data augmentation by factor 100) we perform NNS searches to find correct training labels $\Theta_k$ (and not those originally generated the fingerprints):  
\eql{ 
	\Theta_k:= \text{NNS}_{\widetilde D}(\widetilde D_k+\xi_k),
}
which enables learning a projection mapping rather than a denoiser.


\section{Geometry of deep learning for Bloch response manifold projection}
In this part we show that the MRF-Net provides a piece-wise affine approximation to the Bloch response manifold projection. Our analysis is inspired by the recent work~\cite{splineDN} and is in relation to the authors' previous works~\cite{CoverBLIP-MLSP, coverblip-journal} on adopting cover tree data structures to cluster dictionary atoms in hierarchical segments and using fast approximate NNS search for Dictionary Matching (DM). We show that MRF-Net also clusters the input space, however as opposed to the cover trees, the network does not memorize the MRF dictionary but rather efficiently encodes a set of deep (hierarchical) \emph{matched-filters} for affine regression of the NMR quantities in each segment. Further, segments here have piece-wise affine boundaries.     

\subsection{Affine spline function approximation}
For a network composed of piecewise linear activation functions such as ReLU and the (linear) identity activation, we have that:
{\rem{
Each layer's output $h^{(i)}$ is a piece-wise affine transformation of its direct input $h^{(i-1)}$. Composition of  such layers gives mappings $h^{(i)}(x):\RR^{L}\rightarrow \RR^{\dim(i)}$ which $\forall i$ are piece-wise affine transformations of the input $h^{(0)}=x$ (see e.g.~\cite{montufar2014number}). Further, using continuous activation functions (as above) and for bounded $\{W^{(i)}, \beta^{(i)}, i\}$, we have that $h^{(i)}$  is \emph{Lipschitz continuous}. 
}}

In MRF-NET the last non-linearity is mainly used to impose non-negativity of the estimated parameters, and therefore most of the prediction task is done by the preceding layers. We denote by 
\eql{z^{(N)} (x):= W^{(N)} h^{(N-1)}+\beta^{(N)} }
as the \emph{weighted outputs} before the last non-linearity. 
We have the following \emph{affine spline} representation for the weighted outputs~\cite{splineDN}:
\eql{
	z^{(N)}(x) = \A[x]x+\b[x] :=\sum_r \left( A_r x+b_r \right)\mathbf{1}_{\Omega_r}(x), \label{eq:spline}
	}
where $\mathbf{1}_{\Omega_r}(x)$ is the indicator function with respect to a segment $\Omega_r\in \RR^L$, returning $x$ if it belongs to the segment and 0 otherwise \textemdash segments form a disjoint partitioning of the input space. Matrices $A_r\in\RR^{P\times L}$ and vectors $b_r\in\RR^P$ define the corresponding input-output affine mapping for each segment. We use the shorthands $\A[x], \b[x]$ to represent the input-dependent (piece-wise affine) mapping of $z^{(N)}(x)$: $P$ input-dependent offsets represented by $\b[x]$ and similarly,  an input-dependent $P\times L$ matrix $\A[x]$ where each row represents a \emph{matched-filter} (acting on $x$ and measures the mutual correlation) corresponding to a certain output coordinate $p=1,\ldots,P$. 

Match-filters and offsets here are used for \emph{regressing} the outputs \textemdash not to be confused with the MRF dictionary matching (DM). 
 In other word, during training the network $h^{(N)}(x)$ learns $\{W^{(i)}, \beta^{(i)}\}$ or equivalently $\{\A[x],\b[x]\}$ to provide a continuous and piece-wise affine \emph{approximation} between input and output e.g. the $\text{NNS}_D(.)$ function (Bloch manifold projection) for the MRF reconstruction problem. The universal approximation theorem~\cite{cybenko1989} states that a 2-layer shallow network with large enough units can provide an arbitrarily close approximation to any Borel-measurable function. Deeper networks however are often more favourable in practice to efficiently reduce the number of hidden units~\cite{shallowdeep-bengio}. For certain manifold embedding tasks this has been proven e.g. in~\cite{deepmanifold-basri,coifman-manifoldDN,geometricDL,dongdong_manifold}.

\subsection{Visualizing MRF-Net's segments on Bloch manifold}

Finite-sized networks with bounded weights and biases introduce piecewise affine boundaries for their corresponding segments $\Omega_r$~\cite{montufar2014number}. It is easy to verify that each layer $i$ in \eqref{eq:layers} with aforementioned non-linearities introduces segments with piecewise affine boundaries in its direct input space $h^{(i-1)}\in \RR^{\dim(i-1)}$. Further each segment corresponds to an affine transformation of $h^{(i-1)}$. Therefore, composition of such layers 
results in piecewise affine segments $\Omega_r$ in the input space for the function $z^{(N)} (x)$.

{\rem{Continuity of the mapping $z^{(N)}(x)$ implies that adjacent segments $\Omega_r,\Omega_{r'}$  correspond to distinct $A_r,A_{r'}$. Indeed, if $A_r=A_{r'}$ and the only difference is in the offsets $b_r\neq b_{r'}$, then $\Omega_r,\Omega_{r'}$ won't intersect on boundaries. Therefore they are not adjacent segments unless contradicting the continuity assumption.}}

This remark gives an idea for visualizing the segments as follows: for densely sampled input signals $x$, we compute gradients of the weighted output with respect to the input. The gradients determine input-dependant slopes in affine spline formulation \eqref{eq:spline} i.e. rows of $\A[x]$ at a point $x$ are populated as follows $\forall p=1,2,\ldots,P$:\footnote{For a matrix $X$ we denote by $X_{(p,.)}$ as its $p$-th row. For a vector $x$, $x_p$ denotes its $p$-th element. We also later denote by $\mathbf{e}_p$ the coordinate vector whose $p$-th element is one and zero elsewhere.}  
\begin{align}
	\A[x]_{(p,.)}&=\left(\nabla_x z^{(N)}_{p}(x)\right)^T \nonumber\\  
	& :=\left[ \frac{\partial z^{(N)}_{p}(x)}{\partial x_1}, \frac{\partial z^{(N)}_{p}(x)}{\partial x_2},\ldots,\frac{\partial z^{(N)}_{p}(x)}{\partial x_L} \right]. \label{eq:grad}
\end{align} 
For a given input $x$ the gradients in \eqref{eq:grad} can be efficiently calculated using \emph{back-propagation}. We feed forward $x$ to identify all activations $z^{(i)},h^{(i)}$ and then follow the recursion starting at the vector $y = \mathbf{e}_p$: 
\begin{align} 
&y \leftarrow f_i'(z^{(i-1)}) \odot \left( \left(W^{(i)}\right)^Ty\right) \quad i=N,N-1,\ldots,2.\nonumber\\
&\nabla_x z^{(N)}_{p}(x) = \left(W^{(1)}\right)^Ty,
\end{align}
where $\odot$ denotes  the element-wise vector product, and $f_i'(.)$ denotes the (element-wise) derivative of the activation function. For an identity activation $f_i'(.)$ is an all-one vector and for the ReLU activation it returns one for the positive (direct) input coordinates and zero elsewhere. By vector quantization (e.g. k-means clustering) we cluster regions of $x$ which output distinct slopes $A_r$  and identify $\Omega_r$. We note that for a classification problem~\cite{splineDN} used a similar idea to separately identify each hidden layer's segments and then intersect them in a hierarchical (layer-by-layer) fashion to get $\Omega_r$.

\begin{figure}[t!]
	\centering

		\subfloat
		{\includegraphics[height=3.3cm]{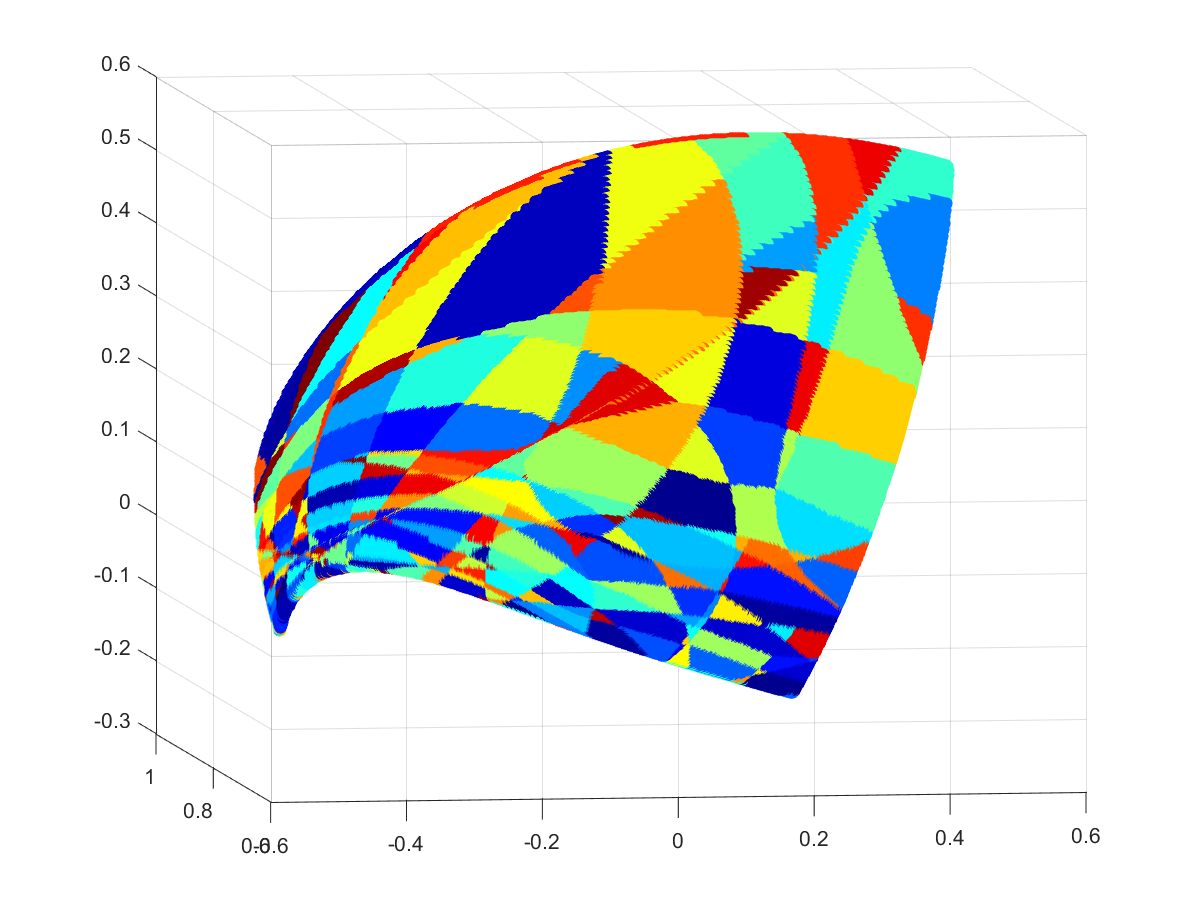} }
		\,\hspace{-.5cm}
		\subfloat
		{\includegraphics[height=3.3cm]{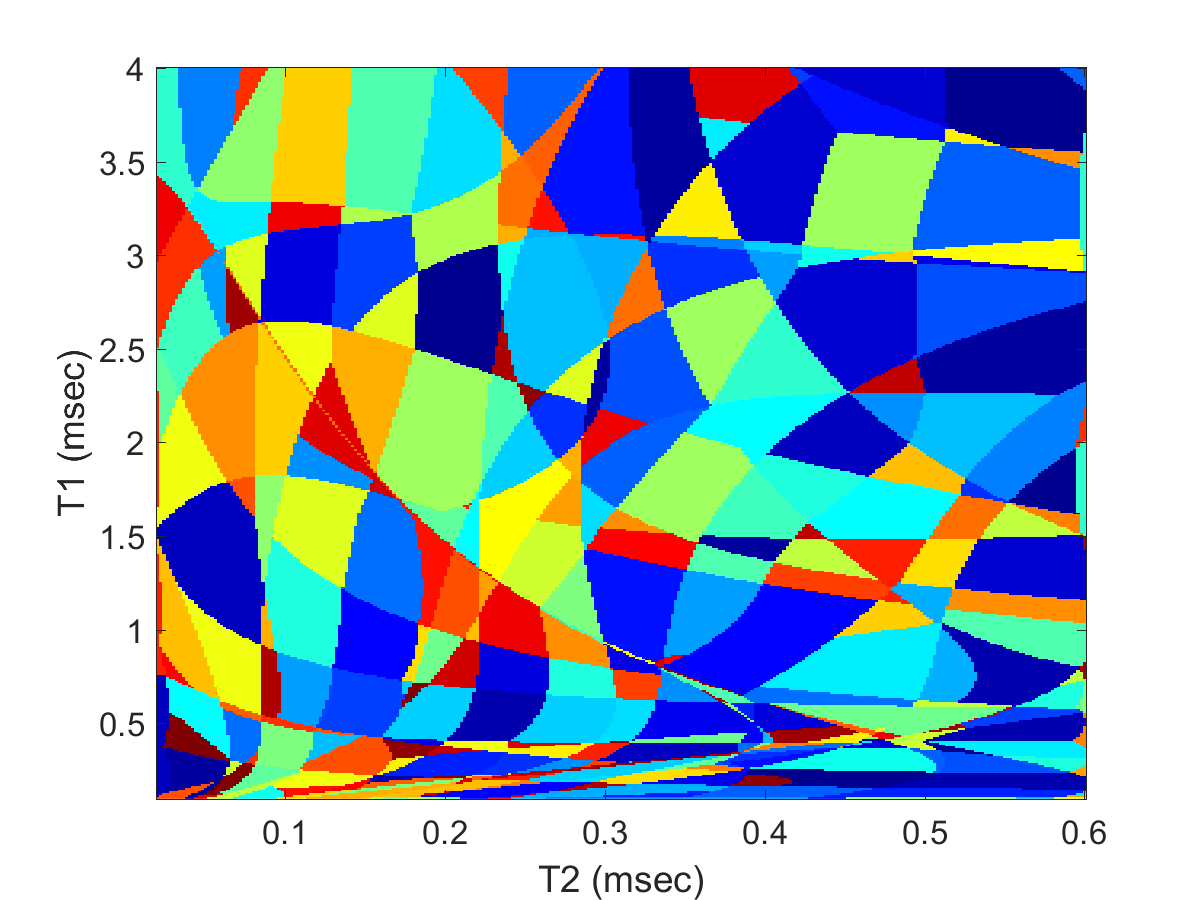} }	
		\caption{MRF-Net's segments on the manifold of Bloch responses to the FISP sequence depicted across the three dominant principal components (left), and 
			the corresponding segments on the $T1,T2$ grid used for generating this dictionary (right).
			\label{fig:seg}}	
\end{figure}

Fine-sampling could be used to visualize intersection of the MRF-Net's segments with the Bloch response manifold. For this purpose we compute \eqref{eq:grad} for inputs corresponding to a dense sample of $T1,T2$ grid i.e. the MRF dictionary. Figure~\ref{fig:seg} visualizes dominant MRF-Net's segments on the manifold of Bloch responses to the FISP sequence used in our experimental validations. 

\subsection{Deep matched-filtering}
\begin{figure}[t!]
	\centering
	
	\subfloat
	{\includegraphics[height=3.3cm]{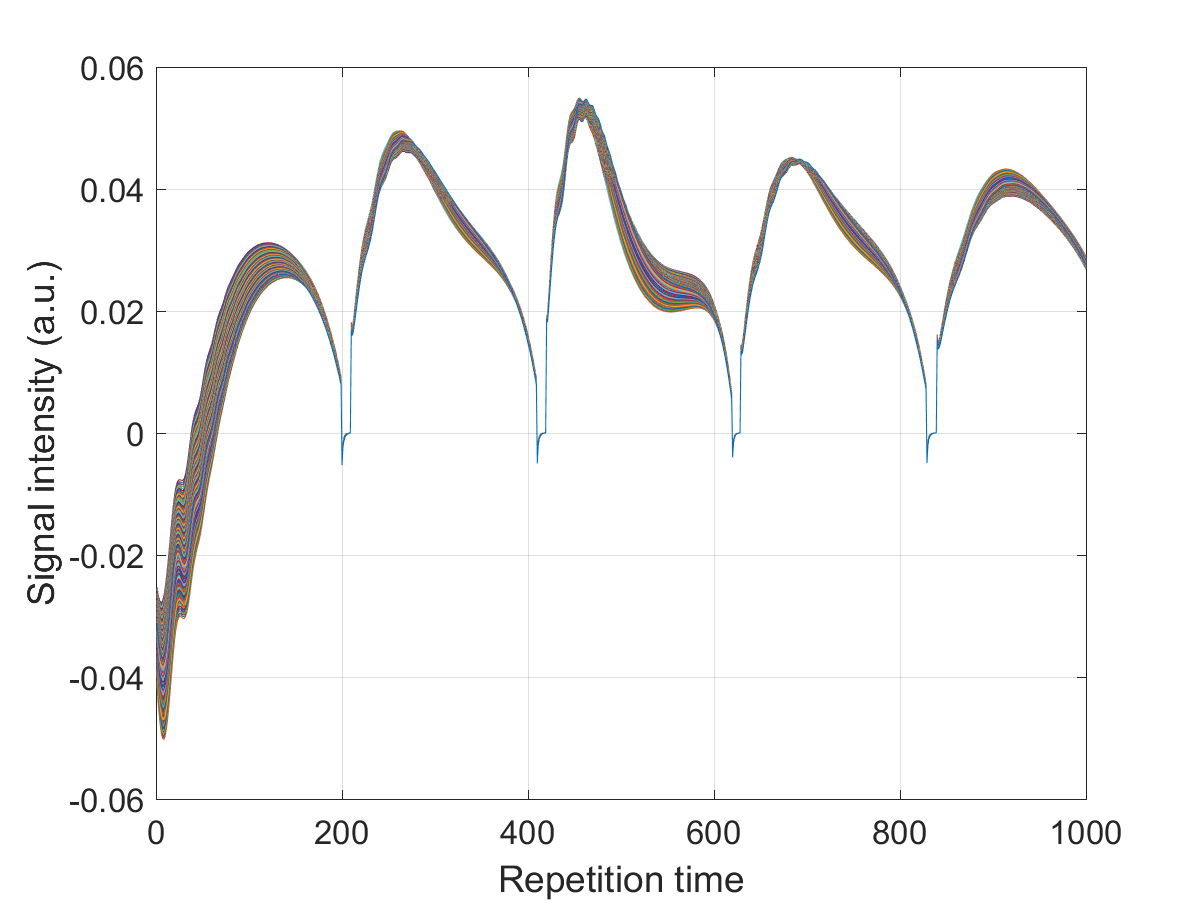} }
	\,\hspace{-.5cm}
	\subfloat
	{\includegraphics[height=3.3cm]{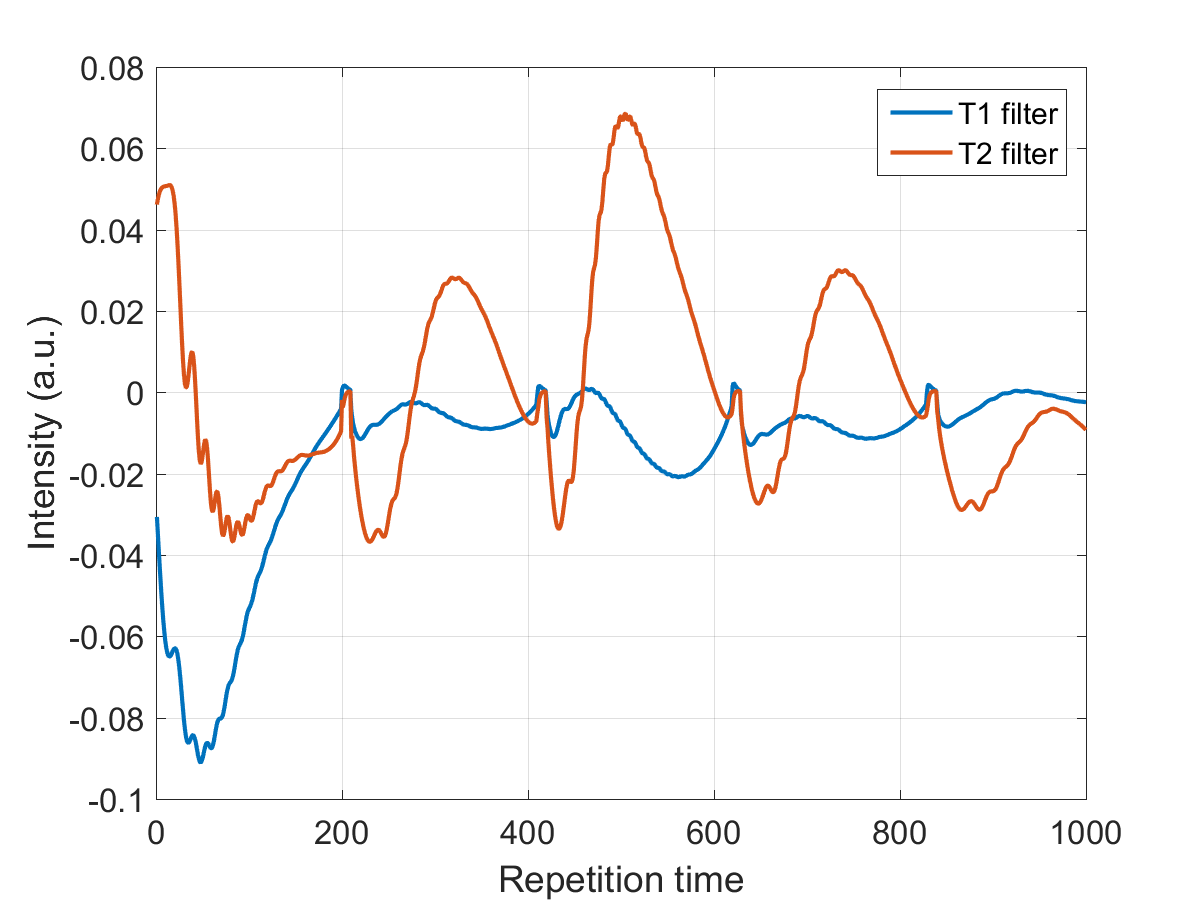} }	
	\\\vspace{-.2cm}
	\subfloat
	{\includegraphics[height=3.3cm]{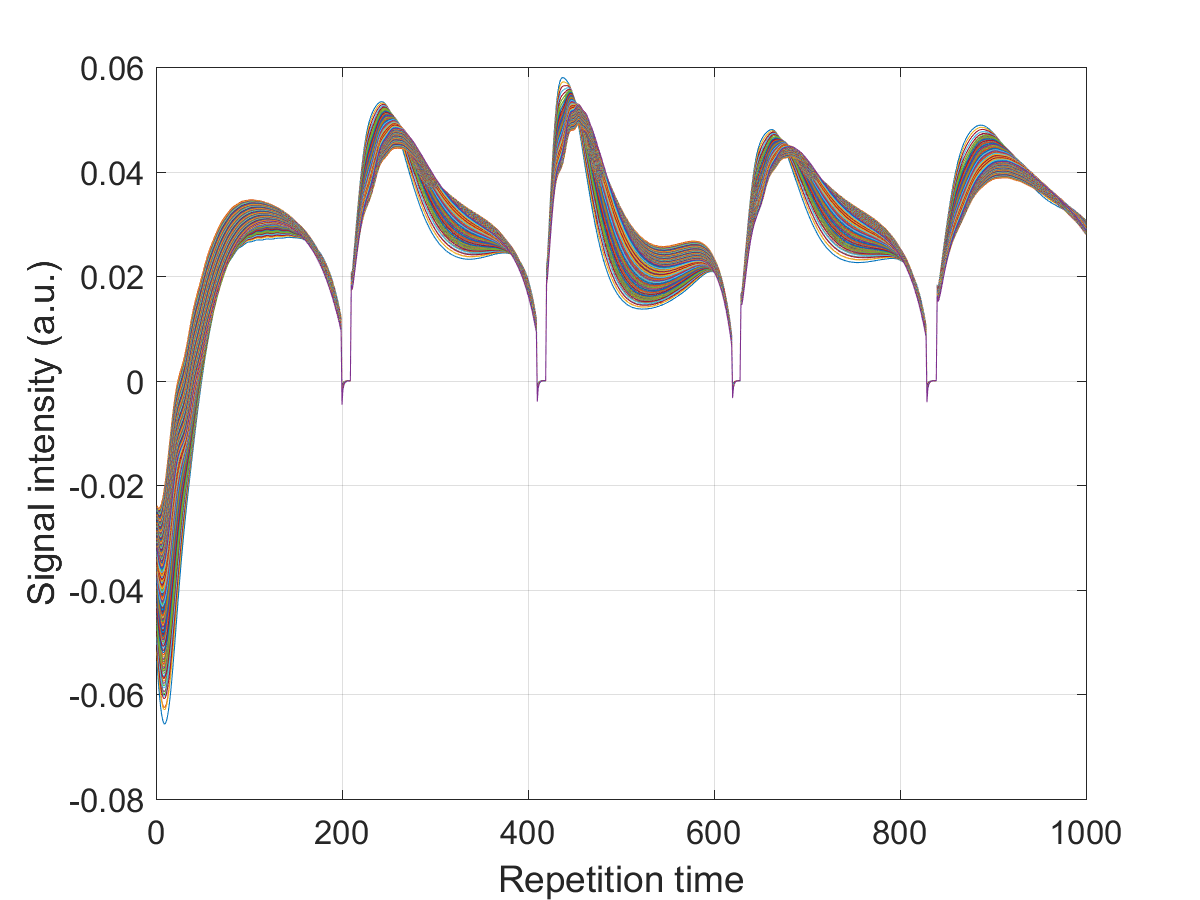} }
	\,\hspace{-.5cm}
	\subfloat
	{\includegraphics[height=3.3cm]{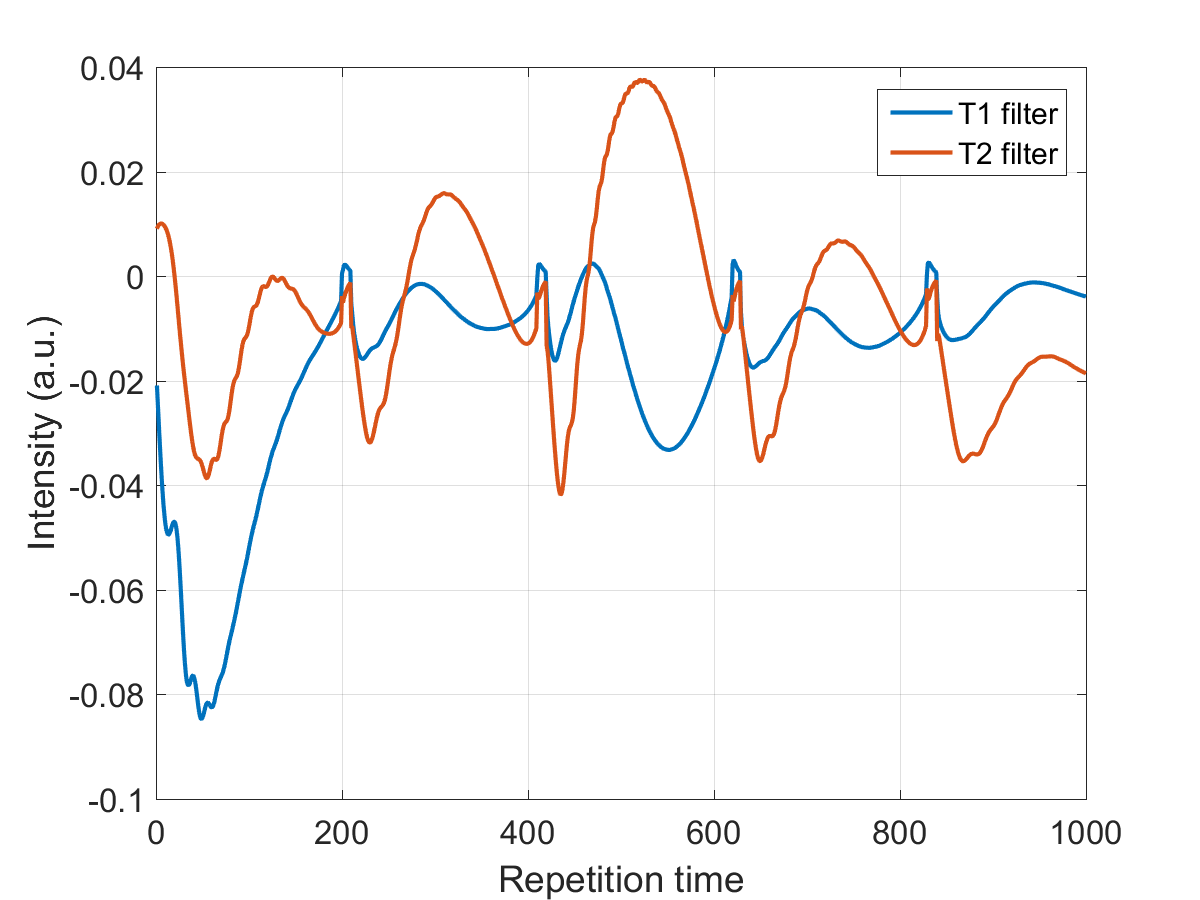} }	
	\caption{FISP dictionary fingerprints (left column) for two segments of the MRF-Net including the standard characteristics measured for Gray (top row) and White (second row) Matters in healthy volunteers' brains. Right column shows the implicit matched-filters that MRF-Net builds for regressing $T1,T2$ quantities in each segment.
		\label{fig:filter}}	
\end{figure}

The identity~\eqref{eq:spline} interprets how data is treated by our network. Ignoring the offsets, the (two) rows of $\A[x]$ correlate with inputs belonging to a segment and predict $T1$ and $T2$. Each segment $\Omega_r$ of the input space has a distinct set of (two) matched-filters i.e. rows of $A_r$, whose correlations with input data (belonging to that segment) will linearly regress the outputs. The end-to-end matched-filtering parameters (i.e. $\{A_r, \Omega_r,b_r\}$) are implicitly learned during training the network and learning the corresponding layer-by-layer affine transformations. In Figure~\ref{fig:filter} we choose two input segments that include standard $T1,T2$ values measured for the White (WM) and Gray (GM) Matters  in healthy volunteers' brains~\cite{t1t2range3T}. We show the clean magnetic responses (i.e. fingerprints) associated with that region together with the matched-filters used for predicting $T1$ and $T2$ quantities. 
As can be seen, \emph{matched-filters peak at discriminant parts of the fingerprints}  that  is where the sequence encodes a significant amount of information (i.e. sensitivity) about the underlying NMR characteristics in that segment. These peaks are visible in the beginning of the FISP sequence due to using an Inversion Recovery (i.e. rotating the external magnetic field by $180^\circ$) and they repeat because of the periodic pattern of flip angles used for FISP excitations (see~\cite[Figure 1.b]{FISP}).

\section{In-vivo experiment}

An \emph{in-vivo} MRF dataset was acquired using the Steady State
Precession (FISP) sequence in~\cite{FISP} and spiral readouts which sample $m=732$ k-space locations in each of the $L=1000$ time-frames in order to reconstruct $n=256\times256$ resolution parametric $T1$ and $T2$  maps. Other scanning parameters are TE$/T_{inv}$=2/18 msec, 8 head-coils, 3 Tesla GE HDx MRI system (GE Medical Systems, Milwaukee, WI), variable density spiral sampling with 89 interleaves, 22.5x22.5cm$^2$ FOV, 256x256 voxel spatial resolution with 5mm slice thickness.
As discussed in Section~\ref{sec:training}, we simulate a dictionary of $d=113781$ atoms which finely samples the grid $T1\in[100, 4000]\, \text{(msec)}, T2\in[20,600]\, \text{(msec)}$. The baseline DM scheme requires direct access to this dictionary for NNS searches whereas the proposed DL approach  only uses it for data augmentation and training the MRF-Net. Figure \ref{fig:maps} compares the reconstructed parametric maps using DM with brute-force searches and the proposed MRF-Net. Note that the computation-memory complexity of  a  dimension-reduced dictionary matching \textemdash without a fast tree search \textemdash is $\mathcal{O}(snL+snd)$ which in this example is more than 60 times higher than the requirements of the (dimension-reduced) MRF-Net. This comparison is on a moderate-size MRF dictionary encoding only two parameters and we expect that for the emerging applications and dictionaries encoding a large number of intrinsic NMR characteristics e.g. $T2^*$, field inhomogeneity, perfusion, diffusion, etc, this gap substantially grows. We leave this direction  for further future investigations.

\section{Conclusion}	
In this paper we study a dictionary-less deep learning approach for the MRF reconstruction problem. Featuring a subspace compression in its first layer, the proposed MRF-Net is compact, easily trained and is capable of achieving comparable estimation accuracy to a dimension-reduced DM baseline, however, with 60 times
less storage and computations. 
The MRF dictionary is only used for training and not during image reconstruction. We show that the MRF-Net provides a piece-wise affine approximation to the Bloch response manifold projection through which, the network efficiently clusters the input space and learns hierarchical \emph{matched-filters} for affine regression of the quantitative parameters in each segment. Future directions could extend this work to applications with a large number of intrinsic NMR characteristics as well as incorporating spatial regularities by e.g. using convolutional networks~\cite{balsigerMRFDL}. 


%
%
\bibliographystyle{IEEEtran}
\bibliography{mybiblio}

\end{document}